\documentclass[journal]{IEEEtran}

\usepackage{amsmath,amssymb,amsfonts}
\usepackage{graphicx}
\usepackage{xcolor}
\usepackage{enumerate}
\usepackage{array}
\usepackage{booktabs}
\usepackage{multirow}
\usepackage{adjustbox}
\usepackage{cite}
\usepackage{listings}
\lstdefinestyle{mypython}{
	language=Python,
	basicstyle=\ttfamily\footnotesize,
	keywordstyle=\color{blue},
	stringstyle=\color{red!70!black},
	commentstyle=\color{green!50!black},
	showstringspaces=false,
	numbers=left,
	numberstyle=\tiny\color{gray},
	frame=single,
	breaklines=true,
	tabsize=4,
}

\begin{document}

\title{Benchmarking and Reasoning Distillation of Large Language Models for Feedback Controller Design in Complex Dynamical Systems}

\author{Zhongchao~Zhou,~Yixuan~Xie,~Wenwei~Yu,~Yuxi~Lu,~Yaonan~Zhu,~Qian~Niu,~Yutaka~Matsuo~and~Yusuke~Iwasawa%
\thanks{This work was supported in part by the JSPS KAKENHI under Grant 26K17341. \textit{(Corresponding authors: Yuxi Lu; Qian Niu)}}
\thanks{Zhongchao Zhou, Yaonan Zhu, Qian Niu, Yutaka Matsuo and Yusuke Iwasawa are with the Matsuo-Iwasawa Laboratory, Department of Technology Management for Innovation, School of Engineering, The University of Tokyo, Tokyo, Japan (e-mail: qian,niu@weblab.t.u-tokyo.ac.jp).}%
\thanks{Yixuan Xie is with the Department of Statistics, University of Illinois Urbana-Champaign, Champaign, IL, USA.}%
\thanks{Wenwei Yu is with the Center for Frontier Medical Engineering, Chiba University, Chiba, Japan.}%
\thanks{Yuxi Lu is with the Shanghai Research Institute for Intelligent Autonomous Systems, Tongji University, Shanghai, China (e-mail: yuxilu@tongji.edu.cn).}}

\markboth{IEEE Transactions on XXX,~Vol.~XX, No.~X, 2026}%
{Zhou \MakeLowercase{\textit{et al.}}: Benchmarking and Reasoning Distillation of LLMs for Feedback Controller Design}

\maketitle

\begin{abstract}
Although remarkable capabilities have been demonstrated by Large Language Models (LLMs) across scientific domains, feedback controller design remains underexplored. Existing benchmarks focus mainly on linear single-Degree-of-Freedom (DoF) systems and large API-hosted models, leaving performance on complex controller-design tasks and feasibility for edge deployment unclear. To address these limitations,  we introduce the Complex Dynamics-to-Control Benchmark for Large Language Models (CoDyControlBench), comprising 132 system configurations across five evaluation dimensions: number of DoF, system type, coupling level, damping regime, and controller type. Six state-of-the-art LLMs were evaluated over three independent runs, including three commercial models (GPT, Gemini, and Claude) and three open-source models (GLM, DeepSeek, and Qwen). GPT achieved the highest overall design success rate at 94.8\%, whereas Qwen showed the lowest rate at 50.0\%. Across the benchmark dimensions, DoF and controller type exhibited the largest model-averaged variations in design success, with success-rate ranges of 36.3\% and 17.6\%, respectively, both exceeding those associated with system type, coupling level, and damping regime. Comparison of GPT and Qwen showed that their performance gap arose mainly from the application of control-design knowledge, particularly gain selection and the use of transient-limiting mechanisms. For edge deployment, a specialized 1.5B-parameter model was developed through reasoning distillation. The reasoning-distilled model outperformed the answer-distilled and base model on CoDyControlBench, maintained stable performance across 1--6 DoFs, and achieved successful traget tracking in all three physical trials on a pneumatic-artificial-muscle-driven robotic arm. These results establish a benchmark baseline and highlight the potential of lightweight, edge-deployable controller-design models.

\end{abstract}


\section{Introduction}

Control engineering underpins a wide range of modern technologies, including robotic automation, mechatronic systems, energy management, and biomedical applications, by enabling stable, accurate, and adaptive system operation \cite{zhu2023shared,kang2026uncertainty}. Among modern control strategies, closed-loop feedback control is especially important because it continuously uses measured system responses to correct tracking errors and maintain robustness under disturbances, uncertainty, and parameter variation \cite{ali2024comparison}. Controller design typically relies on a dynamical model of the plant, obtained through system identification, linearization, or first-principles modeling. When the plant dynamics are represented in state-space form, feedback controller design is formulated as the synthesis of a control law that ensures closed-loop stability and achieves the desired state-regulation or target-tracking performance \cite{ogata2010modern}.

For relatively simple plants, such as linear single-degree-of-freedom (DoF) systems, state-space functions can often be converted into transfer functions in the Laplace domain, enabling classical methods such as root-locus and frequency-response design \cite{kuvcera1999bridge}. These methods provide clear intuition and are therefore central to introductory control education, but they do not fully represent the complexity of real engineering systems \cite{chen2024data}. In practice, multi-DoF coupling, strong nonlinearities, and time-varying parameters make controller design substantially more difficult, requiring control-law expertise, mathematical reasoning, implementation ability, and iterative parameter tuning to maintain stability and performance \cite{nguyen2021robust}.

In recent years, the rapid advancement of large language models (LLMs) has been remarkable. They have demonstrated impressive capabilities in coding \cite{fakhoury2024llm}, reasoning \cite{kojima2022large}, mathematical analysis \cite{frieder2023mathematical}, and domain-specific decision-making \cite{qin2023toolllm}, showing potential to perform tasks that were once considered exclusive to human experts in engineering and scientific domains. This progress has motivated a growing body of work on LLM-assisted controller design and control engineering workflows \cite{de2023llm,guo2024controlagent,narimani2025agenticcontrol,tohma2025smartcontrol,tarczewski2025large,arif2025agentic,guo2025toward,kamenko2025llm,zahedifar2025llm,aydin2026mrac}. Among existing studies, most approaches use natural-language interaction to tune the parameters of an existing controller for a given plant or to design auxiliary compensators that improve reference tracking under predefined operating conditions. However, an increasing number of studies have highlighted two pressing challenges for LLM-based controller design: establishing benchmarks that evaluate LLMs on more complex dynamical systems, and developing lightweight specialized models that reduce reliance on large, cloud-hosted foundation models \cite{zhou2026llm,nosrati2026when}.

The most notable benchmark effort is ControlBench by \cite{kevian2024capabilities}, which introduced a natural-language benchmark for controller design using 147 undergraduate-level control problems drawn from classical textbooks \cite{distefano1997schaum} and control courses at the University of Michigan (EECS 460) and the University of Illinois at Urbana--Champaign (ECE 486). Although ControlBench remains valuable for assessing fundamental control knowledge, its problems are predominantly low-order and linear time-invariant. Consequently, it provides limited discriminatory power for evaluating the controller-design capabilities of current state-of-the-art(SOTA) LLMs.

Regarding model selection, most recent studies rely on large general-purpose foundation models for controller design, including frontier models accessed through APIs, such as GPT-family models, Claude, and Gemini, as well as large open-source LLMs \cite{zhou2026llm,aydin2026mrac}. These studies demonstrate the feasibility of using foundation models as interactive control-design assistants. However, practical deployment remains limited because robotic control systems are commonly implemented on resource-constrained edge devices. Although network connectivity can alleviate local computational constraints by enabling cloud-based inference, it can also introduce communication latency, network instability, bandwidth dependence, data security and privacy risks.

To address these two limitations, we propose the Complex Dynamics-to-Control Benchmark for Large Language Models (CoDyControlBench) to systematically evaluate LLM capabilities in controller design for complex dynamical systems. CoDyControlBench comprises 132 system configurations spanning 1--6 DoFs, four system types, two coupling configurations, and three damping regimes. For each configuration, both heuristic and model-based controller design tasks are evaluated. Six leading LLMs, including three commercial models and three open-source models, are evaluated to assess the current capabilities and limitations of LLM-based controller design. Furthermore, DeepSeek-R1-Distill-Qwen-1.5B is adopted as the base model, and answer and reasoning distillation are performed and compared to develop lightweight control-specialized models. The resulting models are evaluated on CoDyControlBench and further validated through real-world experiments on a one-DoF robotic arm actuated by pneumatic artificial muscle (PAM). The main contributions are summarized as follows:

\begin{enumerate}
	\item We introduce CoDyControlBench, a systematic benchmark for evaluating LLM-based controller design across complex dynamical systems.
	\item We develop a lightweight 1.5B-parameter control-specialized LLM through reasoning distillation for edge-side controller adjustment.
	\item We evaluate six SOTA foundation models and the distilled model on CoDyControlBench, and further validate the distilled model on a real-world one-DoF robotic arm actuated by PAM.
\end{enumerate}

\section{Benchmark Design Principle}

\begin{figure*}[!t]
	\centering
	\includegraphics[width=0.7\linewidth]{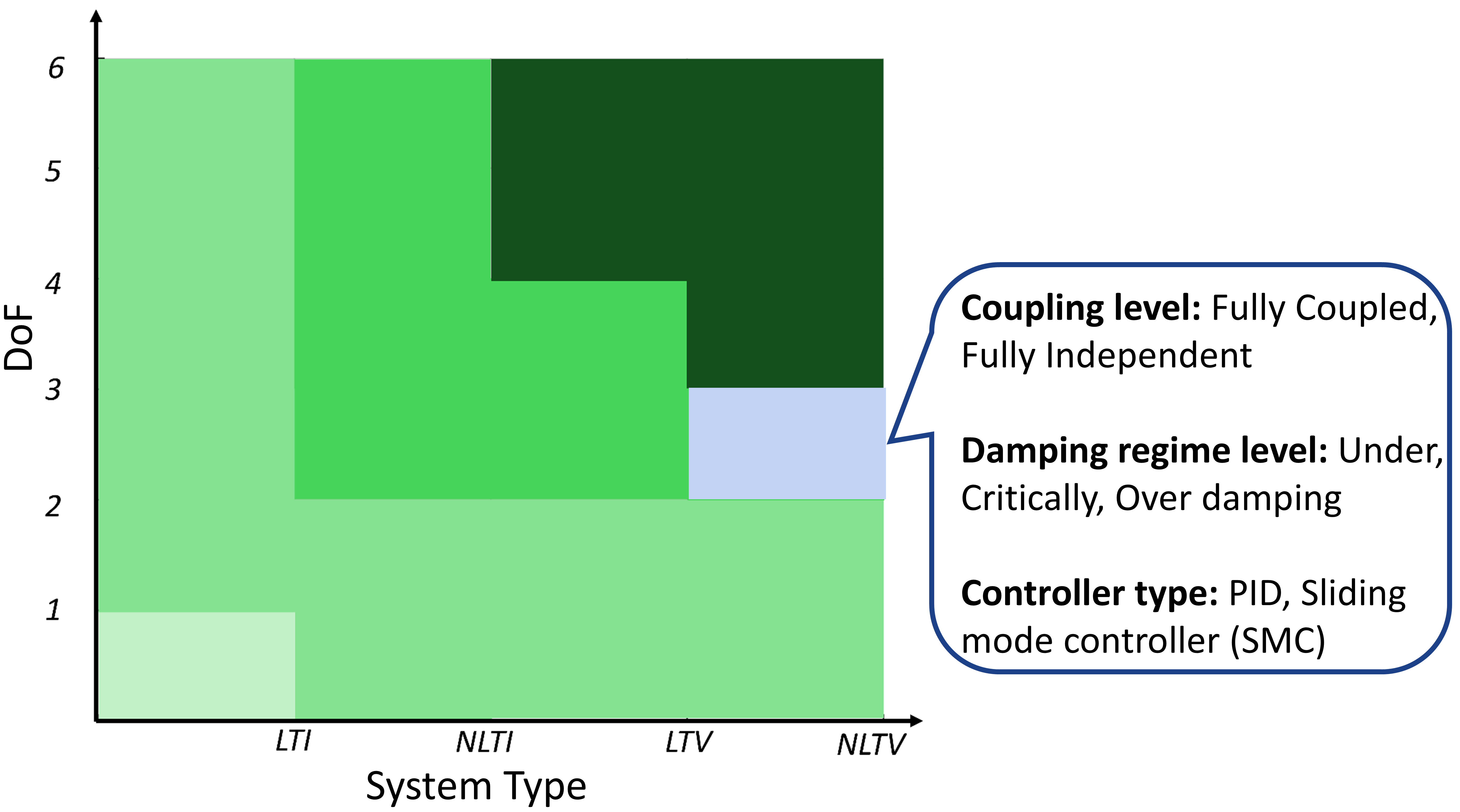}
	\caption{Design principle of CoDyControlBench}
	\label{fig:2}
\end{figure*}

\subsection{Task Taxonomy and Complexity Definition}
To systematically evaluate the capability of LLMs in controller design, as illustrated in Fig.~\ref{fig:2}, CoDyControlBench characterizes controller-design task complexity along five dimensions: \textbf{(1) System Type}, \textbf{(2) Number of DoFs}, \textbf{(3) Coupling Level}, \textbf{(4) Damping Regime}, and \textbf{(5) Controller Type}. The first two dimensions define the primary system structure, while the remaining three capture additional characteristics related to controller-design difficulty.\\
\textbf{(1) System Type:} The system type dimension quantifies the intrinsic dynamic characteristics of the plant based on its governing state-space function. The systems considered in CoDyControlBench are expressed as
\begin{equation}
    \dot{x}(t) = A(x,t)x(t) + Bu(t),
\end{equation}
where $x(t) \in \mathbb{R}^n$ denotes the system state, $u(t) \in \mathbb{R}^m$ the control input, $A(x,t)\in\mathbb{R}^{n\times n}$ is the state-dependent and/or
time-varying system matrix, and $B\in\mathbb{R}^{n\times m}$ is the constant input matrix. The system type is determined by the dependence of $A(x,t)$
on the state $x$ and time $t$. Accordingly, the benchmark systems are categorized into four types:

\begin{itemize}
	\item Linear Time-Invariant (LTI):
	The system matrix $A$ is constant.
	The system therefore exhibits linear and time-invariant dynamics,
	forming the basic setting for classical control design.
	
	\item Nonlinear Time-Invariant (NLTI):
	The system matrix $A(x)$ contains state-dependent nonlinear terms,
	while remaining explicitly invariant with respect to time.
	In this benchmark, the nonlinearities are implemented using
	bounded trigonometric functions.
	
	\item Linear Time-Varying (LTV):
	The system remains linear in the state, while the system matrix
	$A(t)$ varies explicitly with time.
	This category represents systems with prescribed time-varying parameters.
	
	\item Nonlinear Time-Varying (NLTV):
	The system matrix $A(x,t)$ contains both state-dependent nonlinearities
	and time-varying parameters, making it the most complex system type
	considered in this benchmark.
\end{itemize}

This dimension provides a systematic taxonomy of system-type complexity within the benchmark. LTI systems serve as the baseline, whereas NLTV systems combine nonlinearity and time variation and are therefore treated as the most complex system type in this dimension. Between these two extremes, the relative effects of nonlinearity and time variance on design difficulty depend on the system structure and operating conditions \cite{ogata2010modern,nise2019control}. It should be noted that, in the present benchmark, nonlinearity and time variance are introduced in $A(x,t)$ through bounded trigonometric nonlinearities and smooth periodic parameter variations, respectively, while $B$ remains constant to ensure that differences in controller-design difficulty arise from the plant dynamics rather than changes in actuation.

\textbf{(2) Number of DoFs:} The second dimension characterizes the structural scale of the plant. The number of DoFs denotes the number of independent generalized coordinates required to describe the system configuration and determines the scale of its state-space representation. Increasing the number of DoFs enlarges the state and control spaces, increasing the difficulty of system identification, controller design, and stability analysis \cite{zhou2024gan,spong2005robot}. CoDyControlBench considers 1--6 DoFs to cover progressively higher-dimensional dynamics while keeping the benchmark size manageable.

\textbf{(3) Coupling level:} In multi-DoF systems, coupling characterizes the dynamic interactions among different DoFs. To investigate this, two coupling levels are defined:

\begin{itemize}
	\item Fully independent:
	Each DoF is dynamically independent, allowing the system to be decomposed into separable single-input single-output subsystems. 
	
	\item Fully coupled: 
	Neighboring DoFs are dynamically coupled through shared states or torque interactions, resulting in multi-input multi-output dynamics. 
\end{itemize}

Generally, stronger system coupling can increase controller-design complexity \cite{skogestad2005multivariable}. It is important to note that to establish distinct performance boundaries, this benchmark focuses exclusively on the two extremes and excludes intermediate coupling levels (e.g., partial coupling where only two DoFs interact).

\textbf{(4) Damping Regime:}
The damping ratio ($\zeta$) is a continuous dimensionless quantity that characterizes the relative damping level and transient behavior of a dynamical system, whereas a damping regime denotes a categorical range of this quantity used for benchmark classification. To ensure consistent parameterization across the benchmark, a nominal per-DoF damping ratio is defined as
\begin{equation}
	\zeta_i^{\mathrm{nom}}(x,t)
	=
	\frac{c_{ii}(x,t)}
	{2\sqrt{m_i k_{ii}(x,t)}},
	\label{eq:nominal_damping_ratio}
\end{equation}
where $m_i$, $c_{ii}(x,t)$, and $k_{ii}(x,t)$ denote the corresponding diagonal mass, damping, and stiffness terms. For 1-DoF and uncoupled multi-DoF systems, this expression gives the standard damping ratio of each independent coordinate; for fully coupled multi-DoF systems, it defines a coordinate-wise nominal-diagonal indicator based on the effective diagonal entries, rather than an exact modal damping ratio of the complete coupled system. For nonlinear and time-varying systems, the parameters are selected such that $\zeta_i^{\mathrm{nom}}(x,t)$ remains within the prescribed range over the considered operating conditions.

Three damping regimes are defined for every DoF: overdamped ($\zeta_i^{\mathrm{nom}} \ge 2$), critically damped ($\zeta_i^{\mathrm{nom}}\in[0.9,1.1]$), and underdamped ($0<\zeta_i^{\mathrm{nom}}<0.1$). In conventional second-order systems, lower damping ratios are associated with greater overshoot and oscillation, whereas higher ratios suppress oscillations at the cost of slower responses. The selected regimes therefore represent distinct controller-design conditions \cite{dellasantina2021soft}.

\textbf{(5) Controller Type:}
The choice of controller type is critical to the success of control design. In the benchmark, an heuristic error-based controller and a model-based controller are selected:
\begin{itemize}
	\item Proportional-Integral-Derivative (PID)-family: A widely used heuristic, error-based controller that evaluates the LLM's ability to tune control gains using tracking errors and response feedback.
	\item Sliding Mode Controller (SMC)-family: A robust model-based controller that evaluates an LLM's ability to construct sliding surfaces, derive control laws, and analyze stability through mathematical reasoning.
\end{itemize}

The two controller families emphasize different design capabilities. PID-family design primarily evaluates error-driven gain selection and iterative response tuning, whereas SMC-family design emphasizes the construction of sliding variables, reaching laws, and model-dependent control terms. Canonical SMC design generally requires more explicit model-based derivation than PID tuning
\cite{lu2025offline,garcia2019experimental}. 

As summarized in Fig.~\ref{fig:2}, ControlBench in \cite{kevian2024capabilities} is mainly confined to undergraduate-level LTI and 1-DoF configurations, whereas CoDyControlBench extends the evaluation to nonlinear, time-varying, coupled, and high-DoF second-order systems, including NLTV systems with up to 6 DoFs. The resulting benchmark comprises 132 system configurations for evaluating the controller-design capabilities of LLMs.

\subsection{System Prompt Design}
To ensure a fair and rigorous evaluation, a unified system prompt was employed across all experimental queries, the structure and content of the system prompts are detailed in benchmark files.
The specific design requirements and iterative protocols are summarized as follows: Design either a PID or an SMC controller capable of tracking constant step target signals $r_i(t) = 5$ for all DoFs. The simulation horizon is set to $T = 5$s with a sampling interval of $\Delta t = 10$ms. Additionally, three supplementary notes are provided to the LLM to guide the design process:

\noindent\textbf{Note 1:} Variants within each controller family are allowed as long as they remain in that family. The primary objective is to achieve the best possible tracking of each DoF to the constant target \(r_i=5\).

\noindent\textbf{Note 2:} A standardized code template is provided as a structural skeleton. The code template includes the definitions of the controlled system and the two controller types. A brief description can be found in the appendix. The LLM is instructed to fill in the specific [TODOs] within this framework to ensure executability.

\noindent\textbf{Note 3.1:} If Fixed-3 method: the model is forced to perform exactly three iterations, and the successful controller with the highest performance score among the three iterations is used for performance evaluation. If none of the three controllers satisfies the success criteria, the trial is marked as failed.

\noindent\textbf{Note 3.2:} If Success-Stop method: the iteration terminates immediately once the designed controller meets the predefined success criteria. If the design fails, the model is permitted to retry up to the maximum limit of three attempts.

In each step, the feedback (performance metrics or error logs) from the previous iteration is provided to the model, allowing it to progressively refine the controller parameters or structure.

\subsection{User Prompt Introduction}
\begin{figure}
    \centering
    \includegraphics[width=0.8\linewidth]{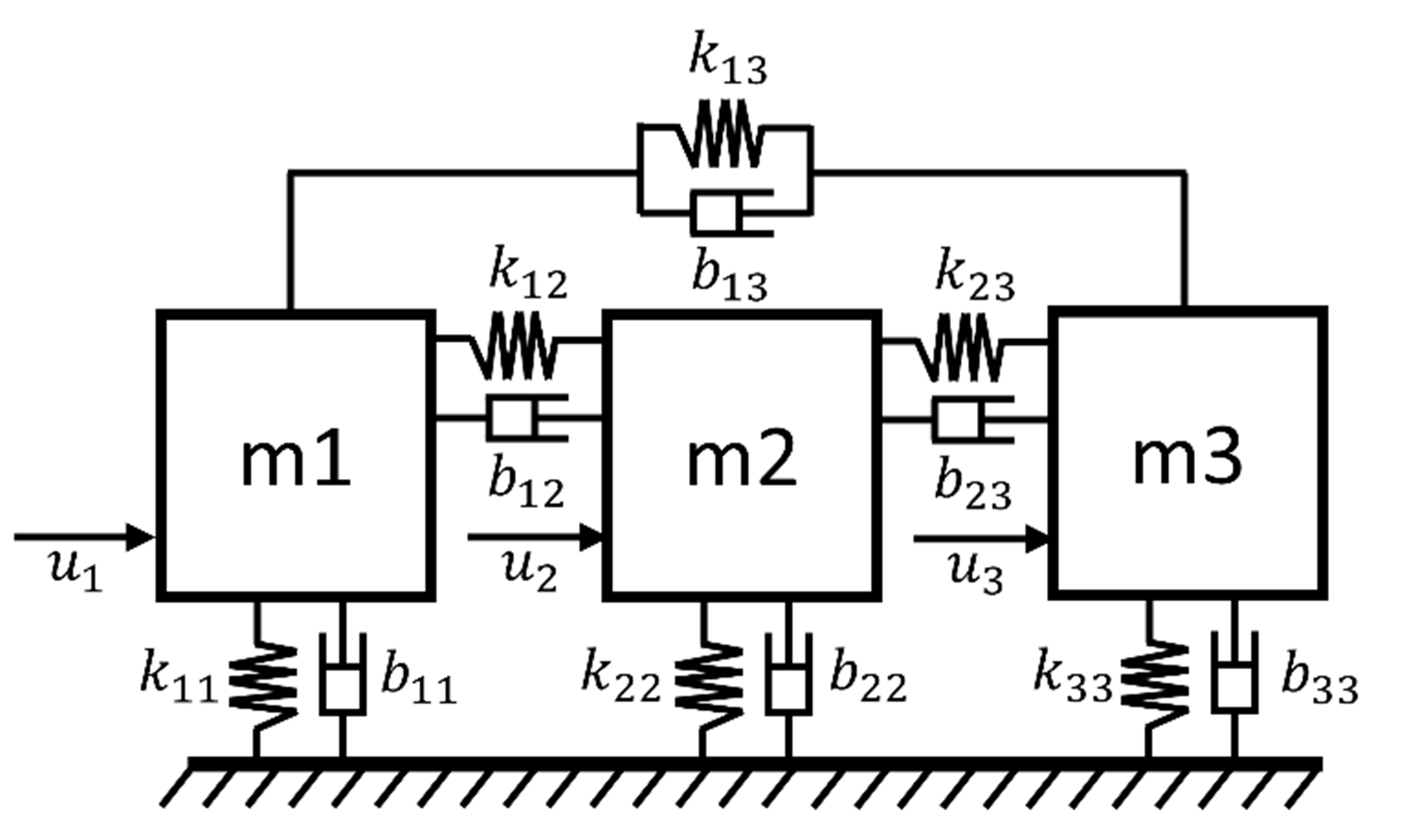}
    \caption{Illustration of Q60. }
    \label{fig:mass-spring-damping}
\end{figure}

For each benchmark case, the question is provided as the user prompt together with the predefined system prompt. To present the benchmark more clearly, Question 60 is selected as an illustrative example. Q60 describes a 3-DoF NLTV underdamped mass--spring--damper system with full coupling among the three DoFs, as illustrated in Fig.~\ref{fig:mass-spring-damping}.
Its state-space function is written as
\begin{equation}
	\dot{x}=A(t,x)x+Bu,
\end{equation}
where
\begingroup
\small
\begin{equation}
\begin{array}{r}
A(t,x)=
\left[
\begin{array}{ccc}
0 & 1 & 0\\
-\dfrac{K_{11}(t,x_1)}{m_1}
&-\dfrac{C_{11}(t,x_1)}{m_1}
&-\dfrac{K_{12}(t,x_1,x_3)}{m_1}\\
0&0&0\\
-\dfrac{K_{21}(t,x_3,x_1)}{m_2}
&-\dfrac{C_{21}(t)}{m_2}
&-\dfrac{K_{22}(t,x_3)}{m_2}\\
0&0&0\\
-\dfrac{K_{31}(t,x_5,x_1)}{m_3}
&-\dfrac{C_{31}(t)}{m_3}
&-\dfrac{K_{32}(t,x_5,x_3)}{m_3}
\end{array}
\right.
\\[3pt]
\left.
\begin{array}{ccc}
0&0&0\\
-\dfrac{C_{12}(t)}{m_1}
&-\dfrac{K_{13}(t,x_1,x_5)}{m_1}
&-\dfrac{C_{13}(t)}{m_1}\\
1&0&0\\
-\dfrac{C_{22}(t,x_3)}{m_2}
&-\dfrac{K_{23}(t,x_3,x_5)}{m_2}
&-\dfrac{C_{23}(t)}{m_2}\\
0&0&1\\
-\dfrac{C_{32}(t)}{m_3}
&-\dfrac{K_{33}(t,x_5)}{m_3}
&-\dfrac{C_{33}(t,x_5)}{m_3}
\end{array}
\right].
\end{array}
\end{equation}
\endgroup

\begin{equation}
	B=
	\begin{bmatrix}
		0&0&0\\
		\dfrac{1}{m_1}&0&0\\
		0&0&0\\
		0&\dfrac{1}{m_2}&0\\
		0&0&0\\
		0&0&\dfrac{1}{m_3}
	\end{bmatrix}.
\end{equation}

The masses and generalized coordinates are
\begin{equation*}
	m_1=m_2=m_3=1,
	\qquad
	q_1=x_1,\quad q_2=x_3,\quad q_3=x_5.
\end{equation*}

The time-varying base stiffness coefficients are
\begin{equation*}
\begin{aligned}
	k_{11}(t)&=3.0+0.4\sin t, \\
	k_{22}(t)&=2.6+0.3\sin(0.8t),\\
	k_{33}(t)&=2.8+0.35\cos(0.6t), \\
	k_{12}(t)=k_{21}(t)&=1.0+0.2\cos t,\\
	k_{13}(t)=k_{31}(t)&=0.8+0.15\sin(0.5t), \\
	k_{23}(t)=k_{32}(t)&=0.9+0.18\cos(1.2t).
\end{aligned}
\end{equation*}

The nonlinear time-varying stiffness entries are defined as
\begin{equation*}
\begin{aligned}
	K_{ii}(t,q_i)
	&=k_{ii}(t)\left(2+\beta_i\sin^2q_i\right),\\
	K_{ij}(t,q_i,q_j)
	&=k_{ij}(t)
	\left[2+\gamma_{ij}\sin^2(q_i-q_j)\right],
	\qquad i\neq j,
\end{aligned}
\end{equation*}
where
\begin{equation*}
	(\beta_1,\beta_2,\beta_3)=(1.2,1.0,1.4),
	\qquad
	(\gamma_{12},\gamma_{13},\gamma_{23})=(1.0,0.8,0.9).
\end{equation*}

The damping entries are
\begin{equation*}
\begin{aligned}
	C_{ii}(t,q_i)
	&=0.18\sqrt{m_iK_{ii}(t,q_i)},
	\qquad i=1,2,3,\\
	C_{12}(t)=C_{21}(t)
	&=0.20+0.05\sin(1.1t),\\
	C_{13}(t)=C_{31}(t)
	&=0.15+0.04\cos(0.8t),\\
	C_{23}(t)=C_{32}(t)
	&=0.18+0.05\sin t.
\end{aligned}
\end{equation*}

The instantaneous nominal per-DoF damping ratios, defined as in Eq.~\eqref{eq:nominal_damping_ratio}, satisfy
\begin{equation}
\zeta_i^{\mathrm{nom}}(t,q_i)
=
\frac{C_{ii}(t,q_i)}
{2\sqrt{m_iK_{ii}(t,q_i)}}
=
0.09<0.1,
\qquad i=1,2,3.
\end{equation}

\begin{figure}
    \centering
    \includegraphics[width=1\linewidth]{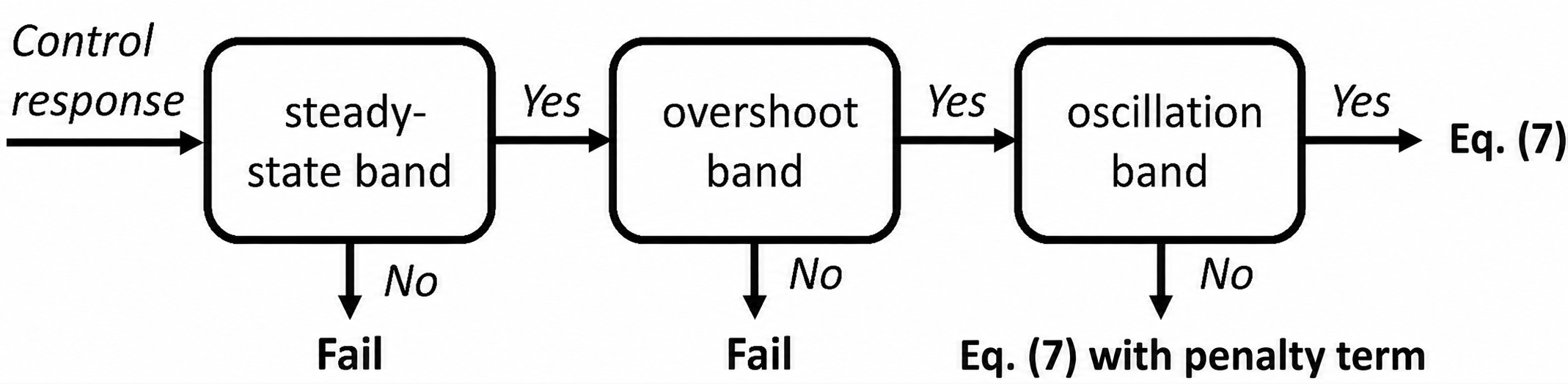}
    \caption{Process of Performance Assessment. }
    \label{fig:Performance Assessment}
\end{figure}

\subsection{Benchmark Assessment and Model Selection}
Evaluating controller performance is inherently challenging, as it depends on multiple dimensions of system behavior. To ensure a consistent and comparable assessment across different control tasks, this benchmark adopts several widely used metrics that jointly reflect steady-state accuracy, transient characteristics, and response speed. \\
(1) Steady-State Error: The steady-state error measures the deviation between the stabilized system output and the target signal. A controller that fails to converge to the target is considered unsuccessful, while a smaller error indicates better tracking accuracy.\\
(2) Overshoot: Overshoot measures the maximum amount by which the output exceeds the reference during the transient response. Excessive overshoot often indicates overly aggressive control gains and undesirable transient behavior; it is therefore heavily penalized.\\
(3) Delay Time: The delay time is the time required for the system output to first reach 50\% of the target. A shorter delay indicates greater responsiveness but may require more aggressive control, reflecting a trade-off between response speed and stability.\\

Considering the interdependence among these metrics, a unified performance score is defined as
\begin{equation}
	\text{Score}
	=
	-\left(
	w_{ess}\lvert e_{ss}\rvert
	+w_{mp}M_p
	+w_{td}T_d
	\right)
	-P_{\mathrm{osc}},
	\label{eq:score_function}
\end{equation}
where \( e_{ss} \) denotes the steady-state error normalized by the absolute target value, \( M_p \) the overshoot normalized by the absolute target value, and \( T_d \) the delay time normalized by the total simulation duration. The \(w_{ess}\), \(w_{mp}\), and \(w_{td}\) are weighting coefficients associated with these three metrics, respectively. A larger weighting coefficient imposes a heavier penalty on the corresponding term. In this work, they are empirically set to \(2\), \(2\), and \(1\), respectively, reflecting a lower tolerance for steady-state error and overshoot. These coefficients can also be adjusted according to practical requirements. The term $P_{\mathrm{osc}}\geq 0$ denotes the terminal-oscillation penalty. Over the final $0.5$~s of simulation, terminal oscillation is identified when the output exhibits at least three effective directional reversals and a normalized peak-to-peak amplitude of at least $0.5\%$. Directional changes smaller than $0.1\%$ of the target magnitude are ignored.  If both criteria are satisfied, $P_{\mathrm{osc}}=5$; otherwise, $P_{\mathrm{osc}}=0$. For multi-DoF systems, the overall score is calculated as the arithmetic mean of the individual DoF scores. A higher (closer-to-zero) score indicates better overall control performance, while more negative scores reflect poorer tracking or instability. 
Before computing Eq.~(\ref{eq:score_function}), each trial is first assessed using predefined pass/fail criteria. A trial that violates either of the following criteria is marked as Fail and excluded from scoring:
\begin{enumerate}[1)]
	\item  During the final 1~s of simulation, all sampled output values are examined. If any sampled point lies outside the $\pm 2\%$ steady-state band, the system is considered not converged and is marked as \textbf{Fail}.  	
    \item 	If the maximum overshoot exceeds a predefined percentage of the target, the run is marked as \textbf{Fail}. In this paper, the threshold is set to \(100\%\), but it can be adjusted according to specific requirements. 	
\end{enumerate}

As illustrated in Fig.~\ref{fig:Performance Assessment}, the response produced by the LLM-designed controller is evaluated against the steady-state band and overshoot requirements. Failure to satisfy either requirement results in a \textbf{Fail}. If both criteria are satisfied, the score is computed using Eq.~(\ref{eq:score_function}), with an additional penalty applied when terminal oscillation is detected. As stated in Note 3.2 of Section~II-B, the Success-Stop criterion is triggered whenever the trial is not marked as \textbf{Fail}. In other words, a trial is counted as successful and the design process is terminated once both pass/fail criteria are satisfied, even if its
performance score is relatively low. For multi-DoF systems, each response channel is assessed independently, and the entire trial is marked as \textbf{Fail} if any DoF violates either pass/fail criterion. These criteria are designed for benchmark-level feasibility screening rather than formal safety verification. The current evaluation focuses on target-tracking accuracy and transient response.

To systematically evaluate controller-design performance across different model paradigms, six leading LLMs were selected. The commercial API models consisted of GPT-5.5 (GPT), Gemini 3.1 Pro (Gemini), and Claude Opus 4.8 (Claude), whereas the open-source models included GLM-5.2 (GLM), DeepSeek-V4-Pro (DeepSeek), and Qwen3.5-397B (Qwen). The abbreviated names in parentheses are used in subsequent tables and discussions. The commercial API models were accessed through their respective official APIs, whereas the open-source models were accessed through OpenRouter. For all models, the reasoning effort was fixed to \texttt{high}, while all other inference parameters followed the default settings of the corresponding API or OpenRouter endpoint. Each benchmark task was evaluated over three independent runs under both the Fixed-3 and Success-Stop protocols. The reported success rates are presented as the mean $\pm$ standard deviation across the three runs, while the performance scores are averaged over successful trials only.

\section{Reasoning Distillation of a Lightweight LLM for Controller Design}
\label{sec:distill}

\subsection{Distillation Strategies}

DeepSeek-R1-Distill-Qwen-1.5B is adopted as the student backbone, motivated by evidence that models at this scale can support local inference on edge-class accelerators following standard deployment optimizations \cite{renney2026cloud}. GPT-5.5 serves as the teacher model and generates the supervision targets. To investigate the role of intermediate diagnostic reasoning, two distillation variants are constructed from matched controller-tuning tasks:

\begin{itemize}
\item \textbf{Answer distillation.} The target consists of a concise rationale followed by executable controller code, without explicit supervision of the intermediate diagnostic process. This setting encourages the student to map the observed response history directly to a controller. The resulting student is denoted as \emph{DeepSeek-1.5B-control-answer} (Answer-Model).

\item \textbf{Reasoning distillation.} The target consists of a teacher-generated reasoning trace followed by the same type of controller code. The reasoning trace describes how the observed system behavior is interpreted and how the corresponding controller modification is selected. The resulting student is denoted as \emph{DeepSeek-1.5B-control-think} (Think-Model).

\end{itemize}

\subsection{Distillation Setup}

A shared set of 599 controller-tuning episodes was generated from randomly sampled LTI systems and used to train both student variants. Each episode contained an unknown plant, a target-tracking objective, up to ten rounds of observed response feedback, and the corresponding controller updates. The system and input matrices were constant within each episode, and the plant parameters were sampled using random seed. No CoDyControlBench instances, response trajectories, or controller solutions were included in the training data. The training systems and CoDyControlBench systems were generated from separately defined parameter distributions, such that evaluation on CoDyControlBench reflects out-of-distribution generalization rather than memorization. The Think-Model and Answer-Model were trained on the same
underlying episodes using identical optimization settings. The Think-Model targets additionally included intermediate diagnostic reasoning traces and were therefore generally longer than the Answer-Model targets; no token-length matching was applied. Both models were initialized from DeepSeek-R1-Distill-Qwen-1.5B and fine-tuned using 4-bit QLoRA with rank $r=16$, scaling factor $\alpha=32$, and dropout 0.05. The LoRA adapters were applied to all attention-projection and feed-forward projection layers. Training used AdamW with a learning rate of $1.5\times10^{-4}$, an effective batch size of 8, a maximum sequence length of 1,792 tokens, and three epochs.

\subsection{Evaluation Protocol}
The base model, the Answer-Model, and the Think-Model are all evaluated on CoDyControlBench. To accommodate their limited model capacity and avoid floor effects, the steady-state band is relaxed from 2\% to 5\%, and the maximum number of tuning iterations is increased from three to ten (Fixed-10). All other settings remain unchanged.  For a fair comparison,  DeepSeek-V4-Pro is additionally evaluated under the same configuration (PID, Fixed-10, 5\%  steady-state band).

In addition to the CoDyControlBench evaluation, the three models are further tested on a real world one-DoF robotic arm driven by PAM. PAM exhibits strong nonlinearity, hysteresis, and pressure-dependent dynamics, making them a challenging real-world platform for evaluating controller design and tuning performance. The dynamic characteristics and dynamic function of the robotic arm can be found in \cite{zhou2022optimization}. During the experiment, the control target is set to $50^\circ$. The joint angle and angular velocity are measured using a BNO055 inertial measurement unit (IMU) and transmitted to an NVIDIA Jetson AGX Orin 64GB Developer Kit through an inter-integrated circuit (I2C) bus. The Jetson executes the distilled model and controller locally and generates the actuator command through a hardware pulse-width-modulation (PWM) output. The Jetson was operated under its default power and clock settings without manual performance tuning. For each target angle, each model is evaluated in three repeated trials.

\begin{table}[!t]
	\centering
	\caption{Overall controller-design performance comparison across different LLMs.}
	\label{tab:total_compare}

	\setlength{\tabcolsep}{4pt}
	\renewcommand{\arraystretch}{1.1}

	\begin{tabular}{lcccc}
		\toprule
		& \multicolumn{2}{c}{Fixed-3} & \multicolumn{2}{c}{Success-Stop} \\
		\cmidrule(lr){2-3} \cmidrule(lr){4-5}
		Model & Succ.\ (\%) & Score & Succ.\ (\%) & Score \\
		\midrule
		GPT
		& 94.8\,$\pm$\,0.8 & -0.60
		& 91.7\,$\pm$\,2.0 & -1.05 \\

		Gemini
		& 90.4\,$\pm$\,1.2 & -0.77
		& 91.4\,$\pm$\,2.5 & -1.66 \\

		GLM
		& 82.3\,$\pm$\,5.0 & -1.03
		& 80.3\,$\pm$\,3.7 & -1.59 \\

		Claude
		& 58.0\,$\pm$\,2.6 & -0.58
		& 58.7\,$\pm$\,1.9 & -0.68 \\

		DeepSeek
		& 55.1\,$\pm$\,2.1 & -0.97
		& 53.3\,$\pm$\,2.3 & -1.23 \\

		Qwen
		& 50.0\,$\pm$\,2.7 & -0.89
		& 46.8\,$\pm$\,2.1 & -1.11 \\
		\bottomrule
	\end{tabular}
\end{table}

\section{Results and Discussion}

\begin{table*}[!t]
	\centering
	\caption{Performance across 1--6 DoFs under the Fixed-3 strategy.}
	\label{tab:DoF_compare}

	\huge
	\setlength{\tabcolsep}{10pt}
	\renewcommand{\arraystretch}{1.1}

	{
	\begin{adjustbox}{max width=\textwidth}
		\begin{tabular}{lcccccc}
			\toprule
			Model & DoF=1 & DoF=2 & DoF=3 & DoF=4 & DoF=5 & DoF=6 \\
			\midrule

			GPT
			& $100.0 \pm 0.0\,(-0.01)$
			& $100.0 \pm 0.0\,(-0.15)$
			& $98.6 \pm 2.4\,(-0.54)$
			& $89.6 \pm 10.4\,(-0.90)$
			& $94.4 \pm 2.4\,(-0.83)$
			& $88.9 \pm 6.4\,(-0.96)$ \\

			Gemini
			& $100.0 \pm 0.0\,(-0.26)$
			& $100.0 \pm 0.0\,(-0.35)$
			& $90.3 \pm 1.2\,(-0.54)$
			& $87.5 \pm 4.2\,(-1.18)$
			& $89.6 \pm 3.6\,(-1.04)$
			& $79.9 \pm 3.2\,(-1.09)$ \\

			GLM
			& $95.8 \pm 4.2\,(-0.44)$
			& $91.7 \pm 5.5\,(-0.80)$
			& $90.3 \pm 3.2\,(-0.78)$
			& $79.9 \pm 4.3\,(-1.38)$
			& $72.9 \pm 5.5\,(-1.41)$
			& $70.1 \pm 12.6\,(-1.28)$ \\

			Claude
			& $90.3 \pm 4.8\,(-0.69)$
			& $91.7 \pm 7.5\,(-0.50)$
			& $73.6 \pm 2.4\,(-0.88)$
			& $38.2 \pm 1.2\,(-0.41)$
			& $35.4 \pm 7.5\,(-0.38)$
			& $34.7 \pm 5.2\,(-0.43)$ \\

			DeepSeek
			& $90.3 \pm 4.8\,(-0.41)$
			& $91.0 \pm 1.2\,(-0.55)$
			& $63.9 \pm 6.7\,(-0.79)$
			& $46.5 \pm 4.3\,(-1.79)$
			& $27.1 \pm 10.4\,(-1.74)$
			& $29.2 \pm 7.2\,(-1.40)$ \\

			Qwen
			& $73.6 \pm 10.5\,(-1.02)$
			& $77.1 \pm 2.1\,(-0.80)$
			& $61.8 \pm 7.9\,(-0.93)$
			& $29.9 \pm 10.7\,(-0.74)$
			& $38.2 \pm 13.2\,(-0.82)$
			& $31.2 \pm 5.5\,(-1.06)$ \\

			\bottomrule
		\end{tabular}
	\end{adjustbox}}
\end{table*}

\subsection{Performance of six state-of-the-art models on CoDyControlBench}
The experimental results for the six LLMs are summarized in Tables~\ref{tab:total_compare}--\ref{tab:controller_compare}. As shown in Table~\ref{tab:total_compare}, GPT achieves the highest overall controller-design success rate (94.8\%), followed by Gemini (90.4\%) and GLM (82.3\%). Claude, DeepSeek, and Qwen achieve substantially lower success rates of approximately 50\%--60\%. Accordingly, GPT, Gemini, and GLM are categorized as high-performing models, whereas Claude, DeepSeek, and Qwen are categorized as low-performing models. A comparison of the two iteration strategies described in Note 3 of Section 2.2 shows broadly comparable success rates, with the main differences occurring in control quality and computational effort. Among successful Success-Stop trials, the best-performing model (GPT) averages 1.51 iterations, while the worst-performing model (Qwen) averages 1.77, indicating only a small difference in iteration count. Since this strategy-level pattern is already captured in Table~\ref{tab:total_compare}, the remaining analyses focus on task-level difficulty factors under the Fixed-3 setting.

\begin{table}[!t]
	\centering
	\caption{Performance across state coupling conditions under the Fixed-3 strategy.}
	\label{tab:couple_compare}
	\setlength{\tabcolsep}{8pt}
	{
	\begin{tabular}{lcccc}
		\toprule
		& \multicolumn{2}{c}{Independent} & \multicolumn{2}{c}{Coupled} \\
		\cmidrule(lr){2-3} \cmidrule(lr){4-5}
		Model & Succ.\ (\%) & Score & Succ.\ (\%) & Score \\
		\midrule
		GPT & 92.8\,$\pm$\,4.2 & -0.44 & 95.8\,$\pm$\,2.5 & -0.88 \\
		Gemini & 91.7\,$\pm$\,1.4 & -0.80 & 87.2\,$\pm$\,1.3 & -0.85 \\
		GLM & 81.9\,$\pm$\,6.3 & -1.01 & 80.0\,$\pm$\,5.8 & -1.20 \\
		Claude & 48.1\,$\pm$\,1.9 & -0.67 & 61.4\,$\pm$\,5.4 & -0.48 \\
		DeepSeek & 55.6\,$\pm$\,1.7 & -1.02 & 47.5\,$\pm$\,3.0 & -1.14 \\
		Qwen & 43.1\,$\pm$\,3.4 & -1.07 & 52.2\,$\pm$\,1.3 & -0.72 \\
		\bottomrule
	\end{tabular}}
\end{table}

As shown in Table~\ref{tab:DoF_compare}, controller-design performance varies substantially across different numbers of DoFs, with several models exhibiting lower success rates as the dimensionality increases. GPT is the only model that maintains consistently high design success, with success rates remaining above 88\% across 1--6 DoFs. By contrast, the success rates of Gemini and GLM decline from 1 DoF to 6 DoFs by 20.1 and 25.7 percentage points, respectively. Further analysis of Gemini at 6 DoFs shows that it achieves $(80.6 \pm 6.4)\%$ on independent systems and $(79.2 \pm 0.0)\%$ on fully coupled systems, a difference of only 1.4 percentage points. These nearly identical rates suggest that, for Gemini, the high-dimensional performance degradation is driven primarily by increasing dimensionality rather than coupling. A broader analysis of coupling effects across models is provided subsequently in Table~\ref{tab:couple_compare}. A more pronounced degradation is observed for Claude, DeepSeek, and Qwen beyond 3 DoFs, with success rates below 40\% at 5--6 DoFs. Among successful trials, control quality also generally deteriorates with increasing dimensionality: the model-averaged score decreases from $-0.47$ at 1 DoF to $-1.04$ at 6 DoFs. Thus, increasing the number of DoFs is associated with lower design feasibility and control quality.

As shown in Table~\ref{tab:couple_compare}, only small aggregate differences are observed between the two coupling levels in both design feasibility and control quality. Across the six LLMs, independent and coupled multi-DoF systems achieve mean success rates of 68.9\% and 70.7\%, respectively. Among successful trials, the mean score decreases slightly from $-0.84$ to $-0.88$, corresponding to an absolute difference of 0.04. Although the direction and magnitude of the differences vary across models, the aggregate results indicate only a small average reduction in control quality under coupling rather than a consistent trend across all models.

As detailed in Table~\ref{tab:system_compare}, LTI systems generally achieve the highest success rates across models. Relative to the LTI case, LTV and NLTI systems generally yield lower success rates, while the NLTV case performs worse than the LTI case for all evaluated LLMs. However, the model-averaged success-rate difference between the LTI and NLTV cases is only 9.3 percentage points, indicating that nonlinearity and time variance tend to reduce design success but have a limited aggregate impact. A similarly limited effect is observed for control quality, with model-averaged scores among successful trials ranging from $-0.74$ to $-0.92$ across the four system types.

\begin{table*}[!t]
	\centering
	\caption{Performance across system types under the Fixed-3 strategy.}
	\label{tab:system_compare}

	\setlength{\tabcolsep}{4pt}
	{
		\begin{tabular}{lcccccccc}
			\toprule
			& \multicolumn{2}{c}{LTI} & \multicolumn{2}{c}{LTV} & \multicolumn{2}{c}{NLTI} & \multicolumn{2}{c}{NLTV} \\
			\cmidrule(lr){2-3} \cmidrule(lr){4-5} \cmidrule(lr){6-7} \cmidrule(lr){8-9}
			Model & Succ.\ (\%) & Score & Succ.\ (\%) & Score & Succ.\ (\%) & Score & Succ.\ (\%) & Score \\
			\midrule
			GPT & 97.0\,$\pm$\,3.0 & -0.55 & 99.0\,$\pm$\,1.7 & -0.73 & 91.4\,$\pm$\,4.9 & -0.51 & 91.9\,$\pm$\,1.7 & -0.62 \\
			Gemini & 93.9\,$\pm$\,3.0 & -0.60 & 90.9\,$\pm$\,0.0 & -0.84 & 86.9\,$\pm$\,2.3 & -0.87 & 89.9\,$\pm$\,1.7 & -0.76 \\
			GLM & 84.8\,$\pm$\,5.5 & -1.11 & 79.8\,$\pm$\,11.5 & -0.95 & 84.3\,$\pm$\,5.7 & -0.80 & 80.3\,$\pm$\,3.0 & -1.29 \\
			Claude & 63.1\,$\pm$\,6.1 & -0.58 & 56.1\,$\pm$\,5.5 & -0.54 & 57.6\,$\pm$\,3.0 & -0.56 & 55.1\,$\pm$\,3.5 & -0.66 \\
			DeepSeek & 63.6\,$\pm$\,7.9 & -0.75 & 57.6\,$\pm$\,5.2 & -1.07 & 55.1\,$\pm$\,5.7 & -0.84 & 43.9\,$\pm$\,4.5 & -1.30 \\
			Qwen & 55.6\,$\pm$\,3.8 & -0.83 & 49.0\,$\pm$\,5.7 & -0.90 & 54.0\,$\pm$\,0.9 & -0.94 & 41.4\,$\pm$\,4.4 & -0.90 \\
			\bottomrule
		\end{tabular}
	}
\end{table*}

\begin{table}[!t]
	\centering
	\caption{Performance across damping regimes under the Fixed-3 strategy.}
	\label{tab:damping_compare}

	\setlength{\tabcolsep}{4pt}
	{
	\begin{tabular}{lcccccc}
		\toprule
		& \multicolumn{2}{c}{Over} & \multicolumn{2}{c}{Critical} & \multicolumn{2}{c}{Under} \\
		\cmidrule(lr){2-3} \cmidrule(lr){4-5} \cmidrule(lr){6-7}
		Model & Succ.\ (\%) & Score & Succ.\ (\%) & Score & Succ.\ (\%) & Score \\
		\midrule
		GPT & 91.3\,$\pm$\,2.9 & -1.00 & 96.6\,$\pm$\,2.0 & -0.43 & 96.6\,$\pm$\,1.1 & -0.40 \\
		Gemini & 87.9\,$\pm$\,4.0 & -0.82 & 91.7\,$\pm$\,4.0 & -0.70 & 91.7\,$\pm$\,3.5 & -0.78 \\
		GLM & 80.7\,$\pm$\,3.9 & -1.11 & 83.0\,$\pm$\,8.2 & -1.01 & 83.3\,$\pm$\,4.7 & -0.98 \\
		Claude & 50.4\,$\pm$\,2.9 & -0.82 & 53.0\,$\pm$\,5.6 & -0.49 & 70.5\,$\pm$\,5.2 & -0.48 \\
		DeepSeek & 48.1\,$\pm$\,1.7 & -1.02 & 57.2\,$\pm$\,4.3 & -0.95 & 59.8\,$\pm$\,3.7 & -0.97 \\
		Qwen & 40.2\,$\pm$\,5.6 & -0.95 & 52.3\,$\pm$\,7.5 & -0.81 & 57.6\,$\pm$\,4.6 & -0.93 \\
		\bottomrule
	\end{tabular}}
\end{table}

The results for the different damping regimes are presented in Table~\ref{tab:damping_compare}. In conventional second-order systems, lower damping ratios are generally associated with greater overshoot and oscillation, often motivating more conservative controller parameters. However, no consistent performance degradation is observed as the damping ratio decreases. Instead, the overdamped regime generally yields lower success rates and poorer scores among successful trials than the critically damped and underdamped regimes. This counterintuitive pattern suggests that the effect of damping is model- and regime-dependent rather than following conventional engineering expectations.

As illustrated in Table~\ref{tab:controller_compare}, a distinct divergence exists between design success rates and performance scores across controller types. The SMC-family condition achieves a higher model-averaged design success rate than the PID-family condition (80.6\% vs.\ 63.0\%), corresponding to a difference of 17.6 percentage points, whereas successful PID-family designs obtain better average performance scores. This result differs from the common expectation that PID-family controllers are generally easier to construct and tune. As discussed in Section~II-A, PID control is traditionally regarded by human practitioners as a more accessible and straightforward methodology. This discrepancy stems from the fundamental difference in their design methodologies. One possible explanation is that the SMC-family template provides a more explicit algebraic structure through the sliding variable and reaching term, which may make controller generation easier for LLMs under limited iterative feedback. In contrast, PID tuning is essentially a heuristic search problem. With only limited iteration-level feedback, LLMs struggle to identify effective gains, leading to a lower overall success rate. However, the superior performance scores of successful PID designs compared to SMC counterparts highlight an inherent trade-off. Once a viable set of PID gains is identified, the controller's linear structure often yields smoother transient responses and lower steady-state errors. Conversely, LLM-generated SMCs frequently prioritize mathematical stability over performance optimization, often resulting in conservative reaching laws or oscillatory output responses that degrade the overall performance score.

\begin{table}[!t]
	\centering
	\caption{Controller-design performance across controller types under the Fixed-3 strategy.}
	\label{tab:controller_compare}

	\setlength{\tabcolsep}{3pt}
	\renewcommand{\arraystretch}{1.1}

	\begin{tabular}{lcccc}
		\toprule
		& \multicolumn{2}{c}{SMC} & \multicolumn{2}{c}{PID} \\
		\cmidrule(lr){2-3} \cmidrule(lr){4-5}
		Model & Succ.\ (\%) & Score & Succ.\ (\%) & Score \\
		\midrule
		GPT
		& 94.9\,$\pm$\,1.2 & -0.63
		& 94.7\,$\pm$\,0.8 & -0.57 \\

		Gemini
		& 94.7\,$\pm$\,2.7 & -1.25
		& 86.1\,$\pm$\,2.9 & -0.24 \\

		GLM
		& 89.6\,$\pm$\,6.1 & -1.57
		& 75.0\,$\pm$\,4.6 & -0.39 \\

		Claude
		& 75.8\,$\pm$\,2.7 & -0.65
		& 40.2\,$\pm$\,5.3 & -0.44 \\

		DeepSeek
		& 62.9\,$\pm$\,3.0 & -1.36
		& 47.2\,$\pm$\,2.4 & -0.46 \\

		Qwen
		& 65.4\,$\pm$\,4.2 & -1.13
		& 34.6\,$\pm$\,1.6 & -0.45 \\
		\bottomrule
	\end{tabular}
\end{table}

\begin{table}[t]
	\centering
	\caption{Descriptive ranges of model-averaged success rates and
	conditional scores across benchmark factor levels.}
	\label{tab:factor_summary}

	\setlength{\tabcolsep}{4pt}
	\renewcommand{\arraystretch}{1.1}
	\footnotesize

	\begin{tabular}{lcc}
		\toprule
		Factor
		& \shortstack{Success-Rate\\Range (\%)}
		& \shortstack{Conditional-Score\\Range} \\
		\midrule
		Number of DoFs  & 36.3 & 0.60 \\
		Controller Type & 17.6 & 0.67 \\
		System Type     &  9.3 & 0.19 \\
		Coupling Level  &  1.8 & 0.04 \\
		Damping Regime  & 10.2 & 0.22 \\
		\bottomrule
	\end{tabular}
\end{table}

As summarized in Table~\ref{tab:factor_summary}, the benchmark factors exhibit different levels of variation in design feasibility and control quality. Their relative variations are summarized by the ranges of the model-averaged success rates and scores across factor levels, with scores computed over successful trials only. The number of DoFs and controller type exhibit the two largest ranges. Increasing dimensionality is associated with lower design success and control quality, whereas controller type introduces a trade-off: model-based controllers achieve higher success rates, while successful heuristic controllers generally obtain better scores. Because model-based design depends on an accurate plant model, which may be uncertain or unavailable in practice, improving LLM capabilities in heuristic, model-free, and robust controller design remains an important direction. The remaining factors exhibit smaller aggregate ranges. NLTV systems yield the lowest aggregate success rate among the
evaluated system types. Coupling has the smallest range for both outcomes, although the observed differences vary across models. Although the overdamped regime is generally more challenging, the aggregate variation across damping regimes remains comparatively limited, without a consistent monotonic trend across the three regimes.

\subsection{Edge-Deployable Distilled Controllers}
Table~\ref{tab:edge_distill} presents the CoDyControlBench performance of the base model, the Answer-Model, and the Think-Model, together with DeepSeek-V4-Pro. Several findings emerge from this comparison. First, the base model exhibits almost no controller-design capability, indicating that the original 1.5B backbone is unable to reliably design effective controllers under the benchmark. Compare to it, the Answer-Model improves over the base model, but the improvement is mainly observed in low-DoF systems. This suggests that answer distillation helps the student learn certain controller-update patterns and improves its ability to generate executable controller code. However, such supervision remains tied to the final artifact and does not generalize well to higher-dimensional dynamics, where its effectiveness quickly diminishes. Moreover, the Think-Model maintains relatively stable performance across 1--6 DoFs, with success rates ranging from 54\% to 69\% and no systematic decline as the number of DoFs increases. The scores also remain consistent across DoFs, indicating that reasoning distillation transfers response-driven tuning strategies that across system dimensions, whereas answer-only distillation does not. The Think-Model therefore demonstrates that dynamics-aware controller adaptation can be retained in a 1.5B-parameter model, supporting the feasibility of lightweight edge-side deployment. Increasing the maximum number of tuning iterations from three to ten substantially improves the success rate of DeepSeek-V4-Pro, particularly for high-DoF systems. At 4--6 DoFs, its average success rate becomes comparable to that of the Think-Model. However, DeepSeek-V4-Pro exhibits substantially larger run-to-run variations at high DoFs, indicating lower consistency across independent runs. In contrast, the Think-Model maintains relatively small variations across these high-dimensional cases, suggesting more reliable controller-design performance.

The real-world results on the PAM-driven robotic arm are presented in Fig.~\ref{fig:pam}, with the corresponding demonstration provided in the supplementary video. Using the 5\% steady-state band, the Think-Model achieved successful tracking in all three repeated trials, corresponding to a success rate of 100\%, whereas the Answer-Model succeeded in only one of three trials, corresponding to a success rate of 33.3\%. Across the three trials, the Think-Model achieved a delay time of $0.239 \pm 0.007$~s, an overshoot of $0.46 \pm 0.40$\%, and a normalized steady-state error of $1.16 \pm 1.38$\%. Its generated controllers rapidly drove the robotic arm toward the target angle and maintained stable tracking with only minor transient fluctuations. The reasoning process of the Think-Model, detailed in Supplementary Material~1, shows an iterative response-driven tuning procedure rather than the direct generation of a fixed controller. A proportional-only controller is first used to probe the unknown plant and infer its input direction, response speed, and residual offset. The proportional gain is then increased, integral action is introduced to reduce the remaining steady-state error, and a small derivative term is finally added while the proportional gain is reduced, yielding a PID-type controller with anti-windup. This process demonstrates that reasoning distillation enables observed system behavior to be translated into structured controller adjustments, explaining the superior and more repeatable real-world performance of the Think-Model. In contrast, the Answer-Model exhibited a comparable delay time of $0.216 \pm 0.005$~s and a small positive overshoot of $0.04 \pm 0.07$\%, but its normalized steady-state error increased substantially to $19.99 \pm 19.24$\%. The small overshoot did not indicate accurate tracking; instead, it resulted primarily from persistent undershooting. Two of the three Answer-Model trials remained below the target and exhibited sustained oscillations, leading to persistent tracking error and a substantially lower experimental success rate.

\begin{table*}[!t]
	\centering
	\caption{CoDyControlBench performance of the 1.5B models and the DeepSeek-V4-Pro}
	\label{tab:edge_distill}

	\setlength{\tabcolsep}{3pt}
	\renewcommand{\arraystretch}{1.1}

	\resizebox{\textwidth}{!}{
	\begin{tabular}{lcccccc}
		\toprule
		Model  & DoF=1 & DoF=2 & DoF=3 & DoF=4 & DoF=5 & DoF=6 \\
		\midrule

		Think-Model
		& $69.4 \pm 3.9\,(-0.60)$
		& $56.9 \pm 10.4\,(-0.81)$
		& $63.9 \pm 5.2\,(-0.71)$
		& $54.2 \pm 5.9\,(-0.69)$
		& $68.1 \pm 7.1\,(-0.69)$
		& $69.4 \pm 5.2\,(-0.67)$ \\

		Answer-Model
		& $50.0 \pm 6.8\,(-0.70)$
		& $30.6 \pm 7.1\,(-0.87)$
		& $0.0 \pm 0.0$\,(\textemdash)
		& $0.0 \pm 0.0$\,(\textemdash)
		& $4.2 \pm 5.9\,(-0.65)$
		& $2.8 \pm 2.0\,(-0.82)$ \\

		Base model
		& 0.0\,(\textemdash)
		& 0.0\,(\textemdash)
		& 0.0\,(\textemdash)
		& 0.0\,(\textemdash)
		& 0.0\,(\textemdash)
		& 0.0\,(\textemdash) \\

		DeepSeek
		& $100.0 \pm 0\,(-0.16)$
		& $100.0 \pm 0.0\,(-0.20)$
		& $80.6 \pm 10.5\,(-0.35)$
		& $68.1 \pm 15.8\,(-0.53)$
		& $65.3 \pm 14.6\,(-0.65)$
		& $66.7 \pm 22.0\,(-0.59)$ \\

		\bottomrule
	\end{tabular}}
\end{table*}

\begin{figure}[!t]
	\centering
	\includegraphics[width=0.8\linewidth]{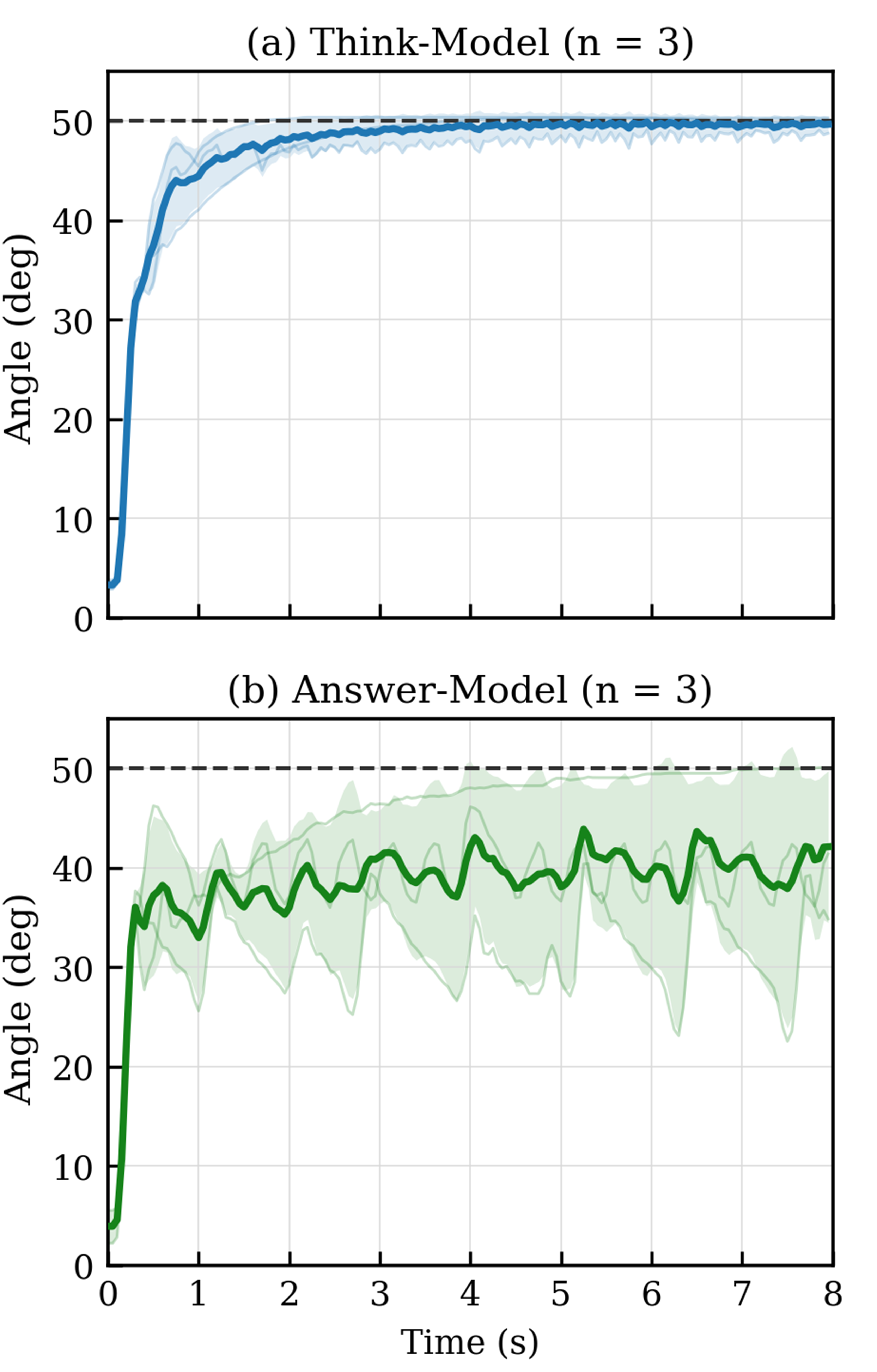}
	\caption{Control responses of (a) the Think-Model and (b) the Answer-Model over three independent trials ($n=3$). 
Dark solid curves show the mean responses, light solid curves show individual trials, shaded regions indicate $\pm 1$ standard deviation, and dashed horizontal lines denote the $50^\circ$ target.}
	\label{fig:pam}
\end{figure}

\section{Detailed Comparison Between the Best and Worst Models}

GPT and Qwen are compared on the high-dimensional subset of CoDyControlBench (DoF $\geq 4$, Cases 61--132) under the Fixed-3 strategy. Their success rates differ substantially, reaching 91.0\% and 33.1\%, respectively. This gap cannot be explained by code-extraction failures, which occur in 0/1296 attempts ($3$ DoF levels $\times$ $24$ systems per level $\times$ $2$ controller families $\times$ $3$ runs $\times$ $3$ iterations) for GPT and only 3/1296 attempts for Qwen. Both models generally produce executable controllers under the PID-family and SMC-family conditions, but they differ substantially in gain selection and the use of transient-limiting mechanisms. For SMC, GPT selects a switching gain 4.2 times larger than that selected by Qwen ($K=65$ versus $15.5$). For PID, its median gains are also substantially larger ($K_p=78$ versus $8$ and $K_i=22$ versus $4$). GPT combines these aggressive gains with output saturation, tight integral limits, conditional integration, and reference shaping. In particular, output saturation is applied in 98\% of its designs, compared with only 21\% of Qwen's designs. Qwen instead manages oscillation primarily by reducing controller gains. This strategy generally preserves stability but prevents accurate convergence. Across 1296 evaluations, only 0.2\% of its controllers diverge, whereas 68.3\% terminate with a steady-state deviation above the required tolerance. Moreover, the initial PID and SMC gains selected by GPT already exceed the corresponding third-iteration gains of Qwen. Thus, under the Fixed-3 strategy, iterative adjustment is insufficient to compensate for Qwen's initially mis-scaled controller parameters.

\section{Conclusion and Future Work}

This work introduced CoDyControlBench, comprising 132 system configurations for evaluating LLM-based feedback controller design across multiple DoFs, system types, coupling levels, damping regimes, and controller types. Six leading LLMs were evaluated over three independent runs. GPT achieved the highest overall design success rate and was the only model to maintain success rates above 88\% across all DoF levels. The number of DoFs and controller type had the strongest effects on both design success and control quality, whereas system type, coupling level, and damping regime showed comparatively limited aggregate effects. NLTV systems generally exhibited lower design success, while the aggregate differences between coupling levels were small overall but varied across models. Although the overdamped regime was generally more challenging, no consistent monotonic performance trend was observed across damping regimes. Model-based design achieved higher success rates, whereas heuristic design generally achieved better control quality among successful trials. The comparison between GPT and Qwen further indicated that their performance gap was associated primarily with the application of control-design knowledge, including gain selection and transient-limiting mechanisms.

Reasoning distillation was further used to develop a lightweight 1.5B-parameter control-oriented model. The resulting Think-Model outperformed the Answer-Model and the base model, maintained stable performance across 1--6 DoFs, and achieved effective reference tracking on a PAM-driven robotic arm. Future work will extend CoDyControlBench beyond the current fully actuated second-order setting by incorporating higher-order and underactuated dynamics, as well as nonlinear and time-varying input mappings in which the input matrix depends on the system state and/or time, i.e., $B=B(x,t)$. Further extensions will consider partial or perturbed model information, measurement noise, external disturbances, actuator constraints, and varying initial conditions and reference signals.

\section*{Conflict of Interest Statement}

The authors declare that the research was conducted in the absence of any commercial or financial relationships that could be construed as a potential conflict of interest.


\appendices
\section{Code Template}

\begin{lstlisting}[
	style=mypython,
	numbers=none,
	breaklines=true
]
import numpy as np
import matplotlib.pyplot as plt

# System definition: specify A and B for x_dot = A x + B u
A = ...
B = ...

# Simulation setup: reference, horizon, and time step
r, T, dt = ..., ..., ...

# Controller hyperparameters (optional)
Kp, Ki, Kd = ..., ..., ...
lam, Ksw, phi = ..., ..., ...

def simulate_pid_family(r, T, dt):
    # 1. Initialize states and buffers
    # 2. Compute tracking error e, derivative de, and integral I
    # 3. Insert PID-family control law
    # 4. Propagate x_dot = A x + B u
    # 5. Record output and control input
    ...
    return ...

def simulate_smc_family(r, T, dt):
    # 1. Initialize states and buffers
    # 2. Compute tracking error e and sliding variable s
    # 3. Insert SMC-family control law
    # 4. Propagate x_dot = A x + B u
    # 5. Record output and control input
    ...
    return ...

if __name__ == "__main__":
    # Run PID-family simulation
    t_pid, y_pid, u_pid = simulate_pid_family(r, T, dt)

    # Run SMC-family simulation
    t_smc, y_smc, u_smc = simulate_smc_family(r, T, dt)

    # Visualize tracking performance
    ...
\end{lstlisting}
\bibliographystyle{IEEEtran}

\bibliography{refs}

\end{document}